# Mechanical Characterization of Compliant Cellular Robots. Part II: Active Strain

**Gaurav Singh**[1]
Department of Mechanical Engineering and Material Science,
Yale University, New Haven, CT, USA
e-mail: gaurav.singh@yale.edu

**Ahsan Nawroj**
Department of Mechanical Engineering and Material Science,
Yale University, New Haven, CT, USA
e-mail: ahsan.nawroj@gmail.com

**Aaron M. Dollar**
Department of Mechanical Engineering and Material Science,
Yale University, New Haven, CT, USA
e-mail: aaron.dollar@yale.edu

## ABSTRACT

*Modular Active Cell Robots (MACROs) is a design approach in which a large number of linear actuators and passive compliant joints are assembled to create an active structure with a repeating unit cell. Such a mesh-like robotic structure can be actuated to achieve large deformation and shape-change. In this two-part paper, we use Finite Element Analysis (FEA) to model the deformation behavior of different MACRO mesh topologies and evaluate their passive and active mechanical characteristics. In part 1, we presented the passive stiffness characteristics of different MACRO meshes. Now, in this part 2 of the paper, we investigate the active strain characteristics of planar MACRO meshes. Using FEA, we quantify and compare the strains generated for the specific choice of MACRO mesh topology and further for the specific choice of actuators actuated in that particular mesh. We simulate a series of actuation modes that are based on the angular orientation of the actuators within the mesh and show that such actuation modes result in deformation that is independent of the size of the mesh. We also show that there exists a subset of such actuation modes that spans the range of deformation behavior. Finally, we compare the actuation effort required to actuate different MACRO meshes and show that the actuation effort is related to the nodal connectivity of the mesh.*

---

[1] Corresponding author.

## 1. INTRODUCTION

Modular Active Cell Robots or MACROs is a design approach for modular robotics hardware [1–3] where the goal is to leverage a large number of simple, identical, and easy to manufacture components to create highly deformable modular robots [4–6]. Using only two types of components, namely simple contracting actuators and passive compliant joints [4] and assembling them in different architectures can result in a range of motion and force behavior. The goal of this paper is to investigate the different mesh architectures that can be used for MACROs and characterize their mechanical behavior. In this two-part paper, we use Finite Element Analysis (FEA) for the mechanical characterization of different MACRO mesh topologies. Based on the physical constraints within the MACRO framework, we narrow down the possible MACRO topologies to the 11 uniform tilings of planar space [7]. In such tilings, the edges represent the linear actuators and the vertices represent the compliant joints. For further details on the choice of MACRO topologies, refer to Part-1 [8]. In Part-1, the *passive stiffness* characteristics of the different mesh topologies are presented and compared. Now, in Part-2 of this paper, we investigate the active characteristics of different MACRO meshes. More specifically, we construct MACRO meshes for the 11 different mesh topologies considered in this paper, followed by actuating certain edges (actuators) within the mesh and evaluating the change in shape (strain) of the complete mesh as a result of this actuation.

MACROs are essentially active lattice structures or shape morphing cellular structures. Internal member actuation of large or infinite lattice structures has been studied by [9–13]. Large cellular truss structures with a single member replaced by an actuator is studied by [9] with a focus on identifying the lattice structure that requires the minimal energy for actuation. Kagome lattice has been studied in detail for shape-morphing lattice structures by [9,10,14], due to its static determinacy and kinematic indeterminacy. Donev et al. [15] show that it is possible to achieve any state of uniform strain by actuating a specific set of bars in an infinite truss. More recently, [11,12] have explored a range of lattice topologies that are suitable for actuated structures. The focus of these papers has been on pin-jointed truss structures with the goal of minimizing the energy required for actuation.

MACROs are similar in concept to the lattice truss structures with the actuators undergoing only linear deformation, identical to the bars in the truss. However, the nodes are molded flexure joints that undergo bending to accommodate the deformation in the MACRO mesh, instead of using pin-joints. The energy stored by these nodes have to be taken into account while modeling the deformation. In addition, although the actuators are significantly stiffer than the nodes in bending, they undergo small bending deformation, especially for topologies with multiple states of self-stress. Due to these characteristics of MACROs and differences with studies in literature [9,11], we use Finite Element Analysis (FEA) to model their behavior. Such a model captures elastic deformation at the nodes as well as the high bending stiffness of the actuators instead of modeling then as perfectly rigid.

In this paper, we also aim to investigate the relationship between the choice of edges in a MACRO mesh that undergo actuation to the overall bulk deformation of the complete mesh. The effect of the choice of edges being actuated within a MACRO mesh on the resulting gross deformation of the mesh is shown in Figure 1. Three different cases of actuation for the Triangular mesh are shown in Figure 1 along with the resulting shape of the mesh upon actuation. In both cases shown by Figure 1(a, b) and (c, d), we actuate the same number of edges. In both cases, 60% of all edges undergo actuation. In Figure 1(a, b), the 60% of the edge undergoing actuation are selected randomly and therefore, we can see that although the overall mesh reduces in size, there is a significant amount of local deformation. Further, in certain cases the contraction of one edge opposes the contraction of another and significant amount of strain energy is stored in the structure requiring more work to be done by the actuators to achieve the deformed shape shown in Figure 1(b). In contrast, Figure 1(c) shows a case where again 60% of the edges undergo actuation. However, here all the edges that are at 0° and 120° with respect to the horizontal are actuated. In this case, the deformation is uniform and there is negligible opposing deformation of edges resulting in much lower strain energy stored in the structure upon actuation and consequently lower work needed to be done by the actuators in deforming the mesh to the shape shown by Figure 1(d). Finally, Figure 1(e) shows the case where all the edges undergo actuation and therefore the deformed shape is trivial and has the bulk strain of the mesh identical to the input strain applied to all edges. Therefore, we can see that the deformation in a MACRO depends on the location and orientation of the edges that undergo actuation within this mesh. A goal of this paper is to study this relationship and quantify the relation between the choice of edges being actuated and the resulting bulk deformation of the MACRO mesh.

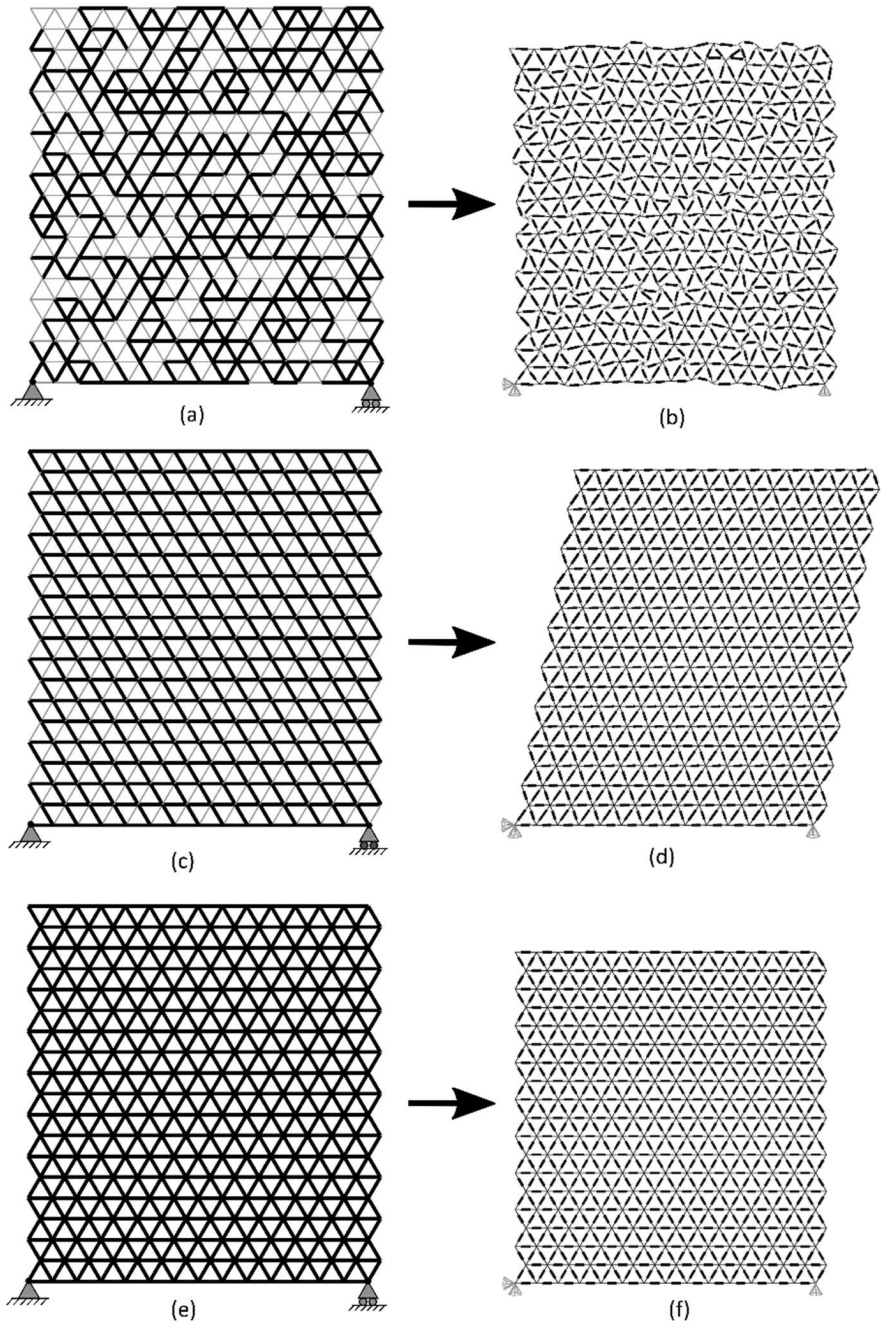

Figure 1. Deformation of Triangular MACRO mesh using FEA. (a) A 28x18 mesh with 60% edges randomly selected to be actuated (Thick edges correspond to the actuated edges (ON) while thin edges correspond to unactuated edges (OFF)) and (b) corresponding deformed shape obtained using FEA. (c) Same triangular mesh but with edges at 0° and 120° (with respect to horizontal measured counter-clockwise) being actuated and (d) corresponding deformed shape. (e) All edges actuated and (f) corresponding deformed shape.

In this paper, we model the deformation behavior of different mesh topologies under the MACRO framework. This model can facilitate the mapping from the applied internal member strain to the resulting overall strain of the MACRO structure. In other words, we quantify the overall mesh strain for the specific choice of the edges being actuated in that mesh. We show that the choice of mesh topology and further the choice of edges that are actuated within this chosen topology determines the strain generated in the whole structure. The main contributions of this work are listed below.

- An FEA model to simulate the internal actuation of different MACRO meshes. This approach can be used to model any active lattice structure with internal member actuation.
- Identification of an actuation scheme that generates deformation that is invariant of the size of the MACRO mesh. This scheme is based on the angular orientation of the edges in the mesh such that the edges that are parallel to each other are actuated together.
- A set of actuation modes that span the complete deformation space, identified through superposition.

## 2. FINITE ELEMENT ANALYSIS (FEA) FRAMEWORK

The goal of this paper is to study the behavior of different MACRO meshes independent of the specific types of actuators or nodes used. Therefore, this study is actuator agnostic and attempts to delineate the mesh topology behavior from the constitutive actuator and node properties. In order to use this FEA framework for simulating a MACRO that uses a particular type of actuator or node would involve measuring their axial and bending stiffnesses, followed by modifying the material properties and cross-section dimensions in the FEA model accordingly.

We create a model where we use the minimum number of parameters to describe the actuators and nodes. We use the commerical package Abaqus to carry out the FEA simulations. For both actuators and nodes, we use Timoshenko beam elements (B21 in Abaqus FEA) with rectangular cross-sections. Since we have used beam elements, the joint effects between actuators and nodes are not modeled in our analysis. Both actuators and nodes are assigned the same material properties but different cross-sectional dimensions. By assigning different cross-sections to these two components, we represent their different stiffness properties. In-plane width of the beams representing actuator is assigned to be 5 mm, while the width of the arms of the flexure nodes are assigned to be 1 mm. We assign the same out of plane width to both the components and it is equal to 5 mm. We assign a Young's modulus of 2000 MPa and Poisson's ratio of 0.3 to both actuators and nodal sections.

Representative meshes of all mesh topologies are generated and then simulated using FEA. To constrain the three rigid body motions in plane, the MACRO mesh is simply supported along the bottom edge as shown in Figure 2, which ensures that the boundary supports do not affect the strains measured by applying the minimal number of boundary supports. Linear strains along $X$ and $Y$ axes, $\varepsilon_x$ and $\varepsilon_y$; and shear strain, $\alpha$ are evaluated for each simulation.

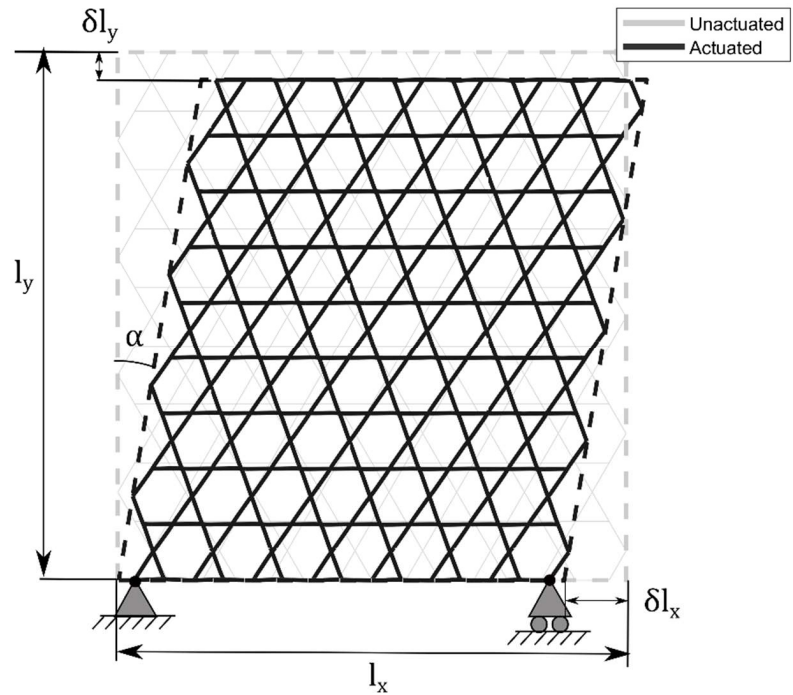

Figure 2. Schematic of a MACRO mesh with boundary conditions used in the FEA model. The mesh is simply supported along the bottom edge to constrain the rigid body motions. To evaluate the strains, a minimal bounding parallelogram is used, shown by dotted lines.

The different steps in the FEA simulation of MACROs to determine the strain characteristics are shown in Figure 3. We start with choosing the mesh topology, size of MACRO mesh being simulated, and the choice of edges within the mesh that are ON, which we refer to as the actuation mode. Using these inputs, a MATLAB function generates the coordinates of nodes, edge connectivity, and a binary variable that stores the edge state (ON or OFF). We import these data in Abaqus to create the part geometry followed by assigning material properties and section. Then, we apply a prescribed contraction to the actuated edges along with the simple support boundary conditions. After solving the FE model, we export data to MATLAB for postprocessing, namely overall strain energy and nodal displacements. To compute strains, we fit the smallest parallelogram that encloses the undeformed and deformed shape of the MACRO. Strains are then evaluated by quantifying the change in size and shape of two parallelograms. We perform these simulations for every possible actuation mode based on the angular orientation of edges for all 11 mesh topologies.

Results of FEA simulations of different mesh topologies are presented in the next section.

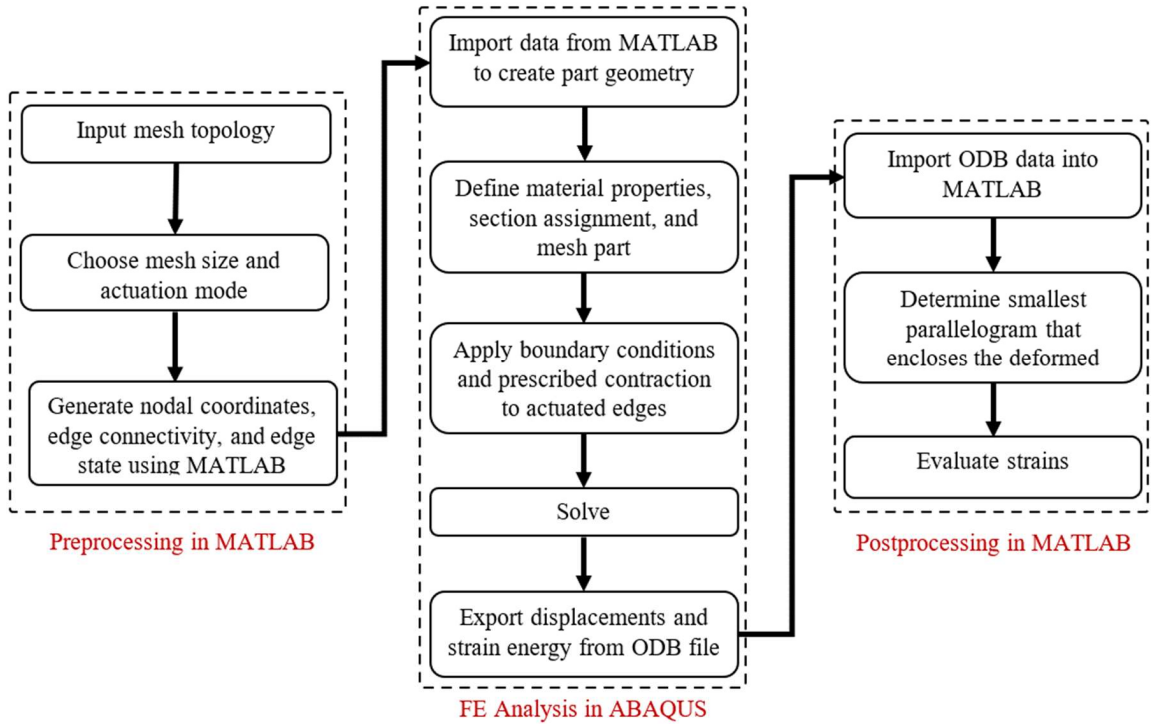

Figure 3. Flow chart showing the step-by-step procedure for evaluating the strains corresponding to different choice of actuated edges of a given MACRO mesh.

## 3. STRAIN CHARACTERIZATION

In this section, we construct meshes of different sizes for all the 11 topologies and then simulate to quantify the overall mesh deformation. By simulating meshes of different dimensions, we aim to obtain strain behavior that is independent of the size of mesh used and instead depends only on the type of mesh topology.

First, we consider the case where all edges of the mesh are actuated. Since, the size of the tiling scales linearly with the length of the edge of the tiling, this case is trivial and the overall mesh strain is equal to the strain in the individual edge. Next, we simulate the case where some but not all edges are actuated. Varying the percentage of ACs being actuated from 20% to 100% with intervals of 20%, the horizontal and vertical strains divided by the applied actuator strain are plotted in Figure 4 for the regular tilings. Here, the actuated edges are chosen randomly for a given percentage of ACs actuated and we run five simulations for each mesh topology and percent of edges actuated. As shown in Figure 4, for 100% actuation overall mesh strain is equal to the applied actuator strain. Also, we can see from Figure 4, for a particular mesh topology and percentage of ACs actuated, we get different amount of strain for different simulations, since for each simulation, the ACs being actuated are picked randomly. Therefore, the choice of cells to be actuated determines the amount of strain of the overall MACRO mesh. Further, it can be shown that for random actuation, the strain also depends on the specific size of the mesh being simulated. Since, we are interested in obtaining MACRO characteristics that are independent of the size of the mesh, we choose an actuation scheme that is based on the angular orientation of the edges. We hypothesize that such an actuation scheme would result in deformation that is independent of the mesh size. We describe this in the next subsection.

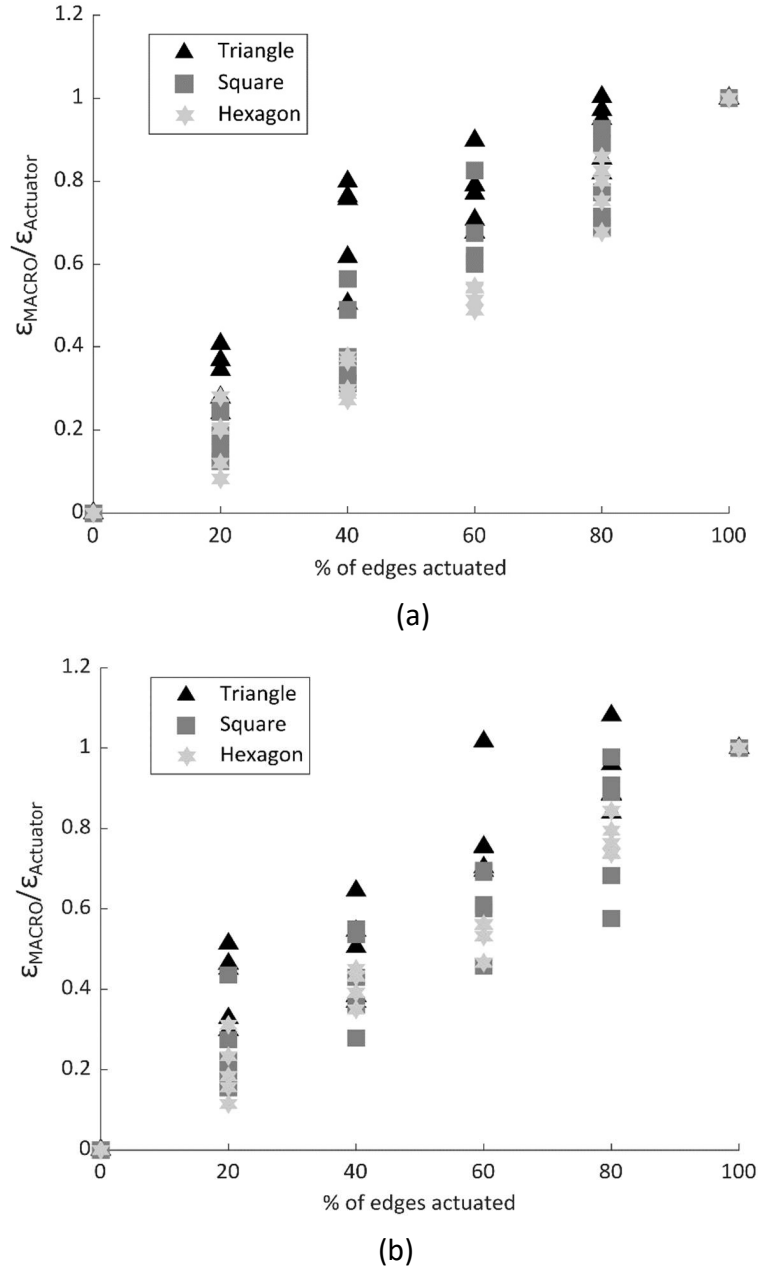

(a)

(b)

Figure 4. Ratio of overall mesh strain to applied actuator strain for different percentages of edges actuated in the MACRO mesh measured along (a) horizontal direction (X axis) and (b) vertical direction (Y axis).

### 3.1 Angular orientation-based actuation scheme

For a MACRO mesh with $n$ actuators that can each be individually controlled to be ON or OFF (if the actuator is ON, it contracts by a specific amount and if it is OFF, it stays in the undeformed state), there are $2^n$ ways in which the mesh can be actuated. Simulating and characterizing the behavior of each of those actuation modes across all topologies is not feasible, but it is also not necessary as there is a huge amount of redundancy in the structure. Therefore, we focus here on an actuation scheme based on the angular orientation of the edges, where all edges that are at the same angular position or in other words that are parallel to each other are controlled together, so they are either all simultaneously ON or all OFF. This choice is based on the assumption that actuators that are parallel to each other would result in similar local deformation and these local deformations would add up towards the overall deformation of the mesh instead of cancelling each other. Further, it can be shown that such an actuation scheme would result in a deformation behavior that is independent of the size of the mesh by analysing meshes of different dimensions and comparing the resulting deformation.

For a mesh topology that has $p$ unique angular orientation of edges, if all the edges with the same angular orientation are controlled together, then there are $2^p$ different modes in which such a mesh can be actuated. Out of these $2^p$ modes, two are trivial corresponding to all edges being OFF (resting configuration) and the other with all being ON, which we have already considered. Here, we simulate the remaining $2^p - 2$ actuation modes. Actuation modes of two of the mesh topologies are shown in Figure 5 and Table 1 lists the number of possible modes of actuation for each type of mesh topology under this angular orientation-based actuation scheme.

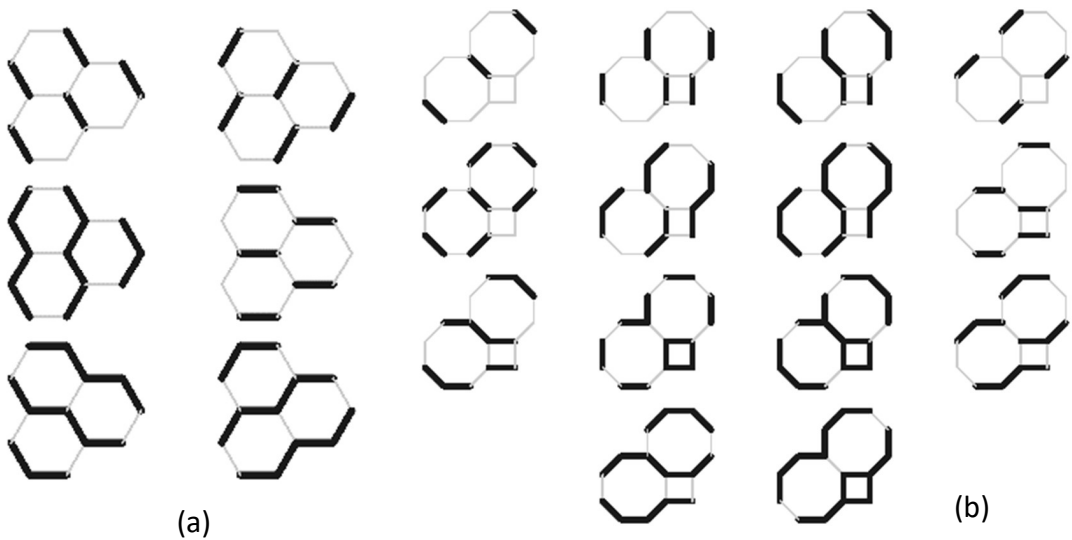

Figure 5. All possible actuation modes for (a) hexagon (H) and (b) SO₂ topologies based on the angular orientation-based actuation scheme. Thick edges correspond to the actuated edges (ON) and thin edges correspond to unactuated edges (OFF).

Table 1. Number of distinct angular orientations for edges and corresponding actuation modes for all mesh topologies

| Topology | Angular orientations | Actuation modes |
|:---:|:---:|:---:|
| S | 2 | 2 |
| T | 3 | 6 |
| H | 3 | 6 |
| THTH | 3 | 6 |
| SHD | 6 | 62 |
| $SO_2$ | 4 | 14 |
| TSHS | 6 | 62 |
| $TD_2$ | 6 | 62 |
| T2STS | 6 | 62 |
| $T_3S_2$ | 4 | 14 |
| $T_4H$ | 3 | 6 |

Linear strains along the horizontal and vertical directions as well as shear strain divided by the applied AC strain are plotted in Figure 6 for each actuation modes of the three regular tilings of the plane. Square (S) topology has only two actuation modes, one where all horizontal edges are actuated and the other where all the vertical edges are actuated. In the first case, horizontal strain ($X$ strain) is equal to the actuator strain and the same is true for vertical strain ($Y$ strain) in the second case. Next, both triangle (T) and hexagon (H) type meshes have six actuation modes each and the corresponding strains are shown in Figure 6(b-c).

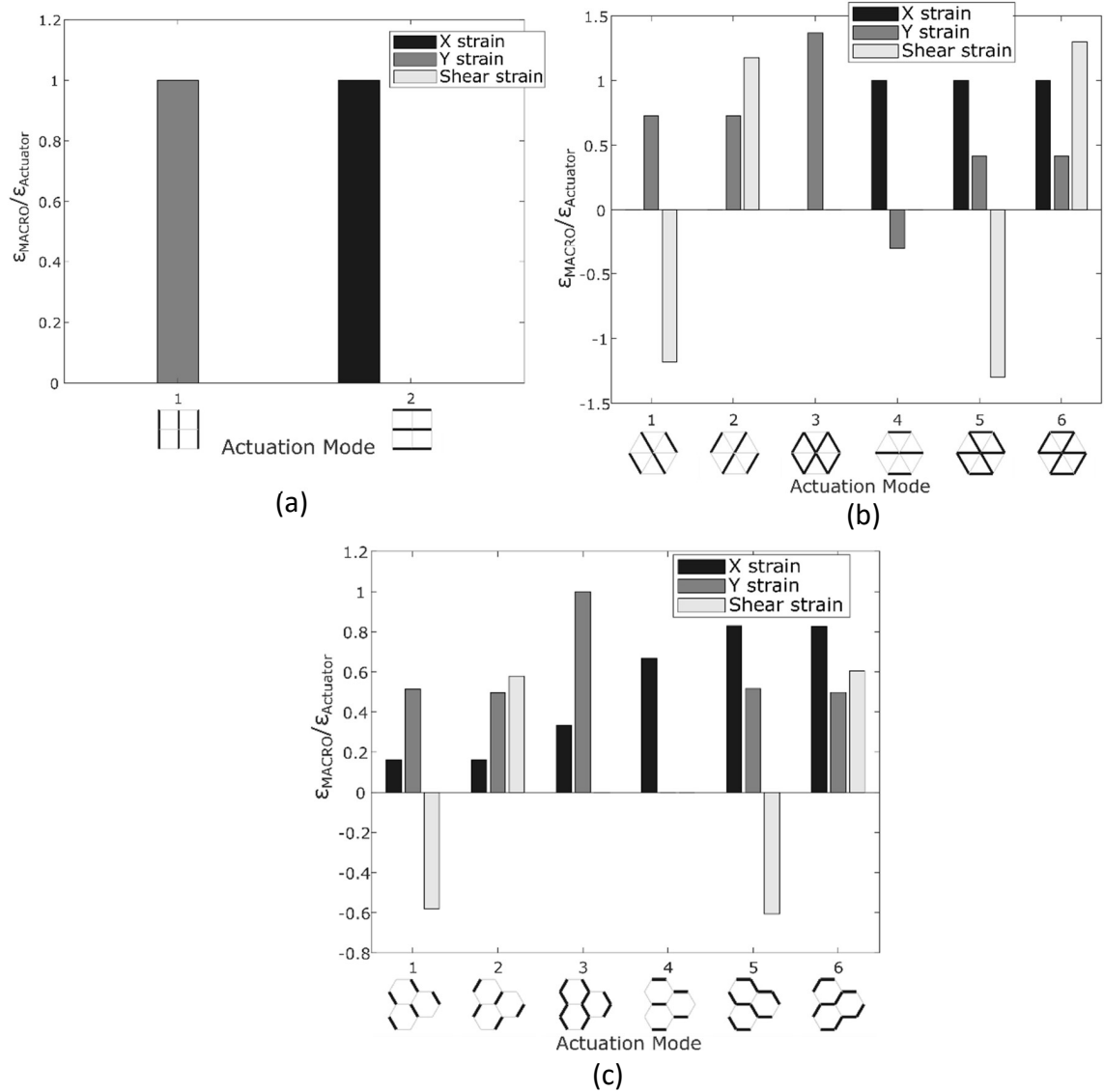

Figure 6. Ratio of MACRO mesh strain to applied actuator strain corresponding to each actuation mode for (a) Square (S), (b) Triangle (T), and (c) Hexagon (H) topologies. The X-annotations denote the respective actuation modes, where edges that are thicker correspond to actuators that are ON and the remaining edges (thin) correspond to actuators being OFF.

The results shown in Figure 6 are invariant of the size of the mesh being simulated, which we have verified by running these simulations on MACRO meshes of different sizes. Further, these results are also invariant of the input actuator strain. Although, results shown here are from the FEA simulations of MACRO meshes with contracting type actuators, the ratio of MACRO strains to input actuator strain are the same for extending type actuators as well.

MACRO to actuator strain ratios vs actuation modes plots for THTH and $T_4H$ type meshes are identical to that of the T type mesh shown in Figure 6(b). Both of these mesh topologies can be obtained from T mesh by selectively removing some of the edges. Therefore, they are simply sparser versions of the T mesh and have the same actuation modes and corresponding strain behaviors as the T mesh.

### 3.2 Spanning Set of Actuation Modes and Superposition

The square (S) mesh has only two actuation modes and if both of these are ON, then all the edges or actuators in the mesh are ON. This results in a deformation where overall mesh strain along both $X$ and $Y$ axis is simply equal to the individual actuator strain and each square in the mesh is reduced in size by the same amount, which is equal to the input actuator strain. In contrast to S's two modes, triangle (T) and hexagon (H) have six actuation modes each. Out of these six, modes 1, 2, and 4 form a spanning set of all the other deformation modes. Strains generated by the remaining actuation modes can be calculated by adding the strains of these three modes. For example, for the T mesh, the strains corresponding to mode 3 can be obtained by adding the strains of modes 1 and 2. This is because the set of edges that are ON in mode 3 is simply the union of the set of edges that are ON in modes 1 and 2. Similarly, adding the strains of modes 1 and 4 gives the strains for mode 5; and adding 2 and 4 gives the strains of mode 6. The same relations hold true for H mesh as well, where modes 1, 2, and 4 span the set of all possible deformations under our actuation scheme. A comparison of strains obtained by this superposition method with those obtained using FEA is shown in Figure 7, where we can see that they match closely. The small error in strain prediction here can be attributed to the bending deformations of the nodes being more accurately captured in FEA as compared to the superposition method.

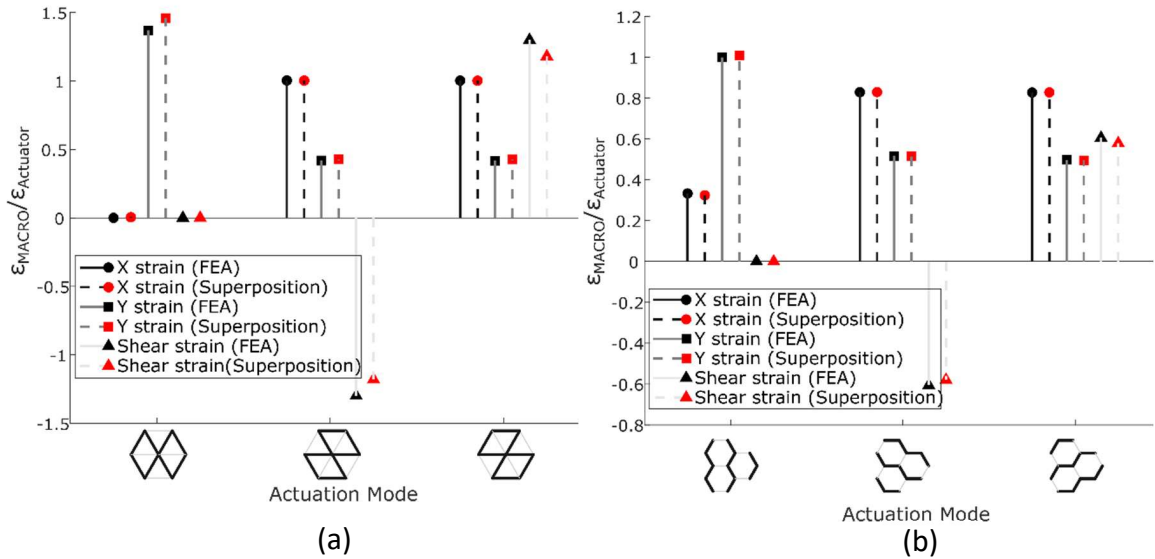

(a)    (b)

Figure 7. Comparison of strains obtained from FEA with the strains obtained by adding the base deformations for (a) Triangle and (b) Hexagon mesh.

Similarly, we can characterize all the actuation modes for the remaining topologies and further identify a subset of these modes that span the whole set. Topologies $T_3S_2$ and $SO_2$, each have 14 actuation modes and four out of those 14 form the spanning set and are shown in

Figure 8. Also, as shown in Figure 9, superposition of these four modes can capture the strain behavior of all the other modes. The remaining four mesh topologies have 64 actuation modes each, and six out of those 64 span the whole set of deformations. This spanning set of deformation modes for these four mesh topologies are shown in Figure 10.

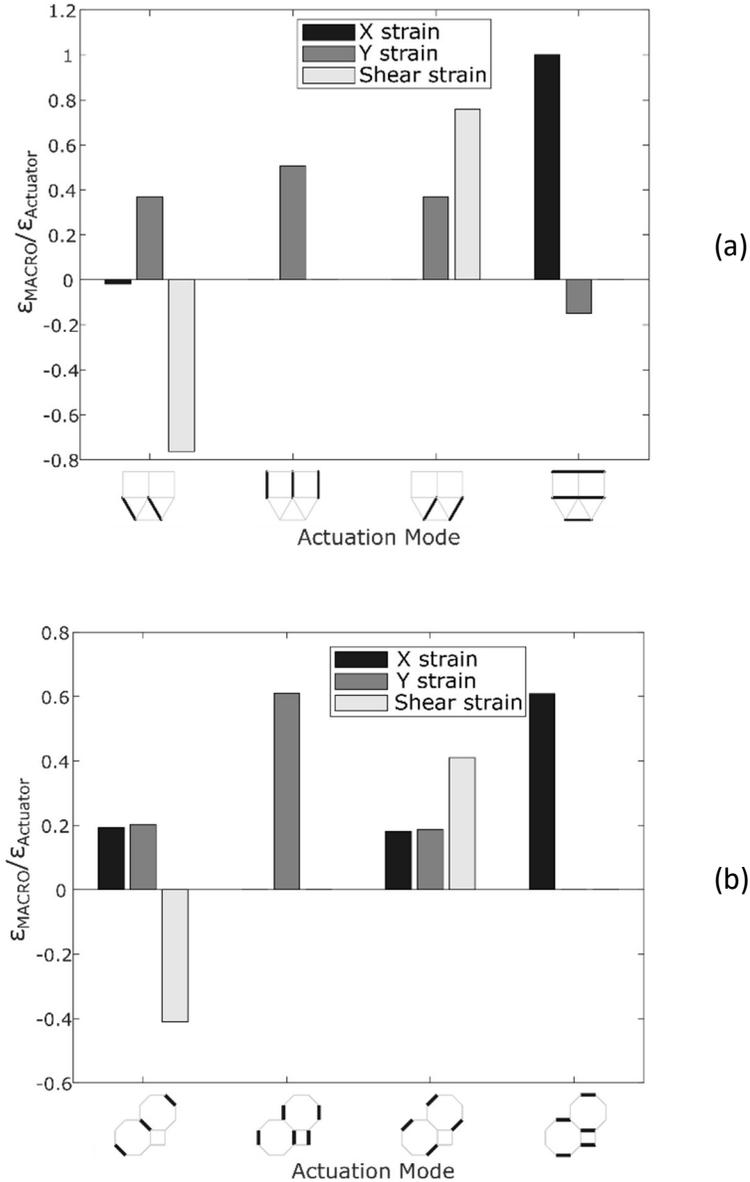

Figure 8. Spanning set of actuation modes of (a) $T_3S_2$ and (b) $SO_2$ mesh topologies.

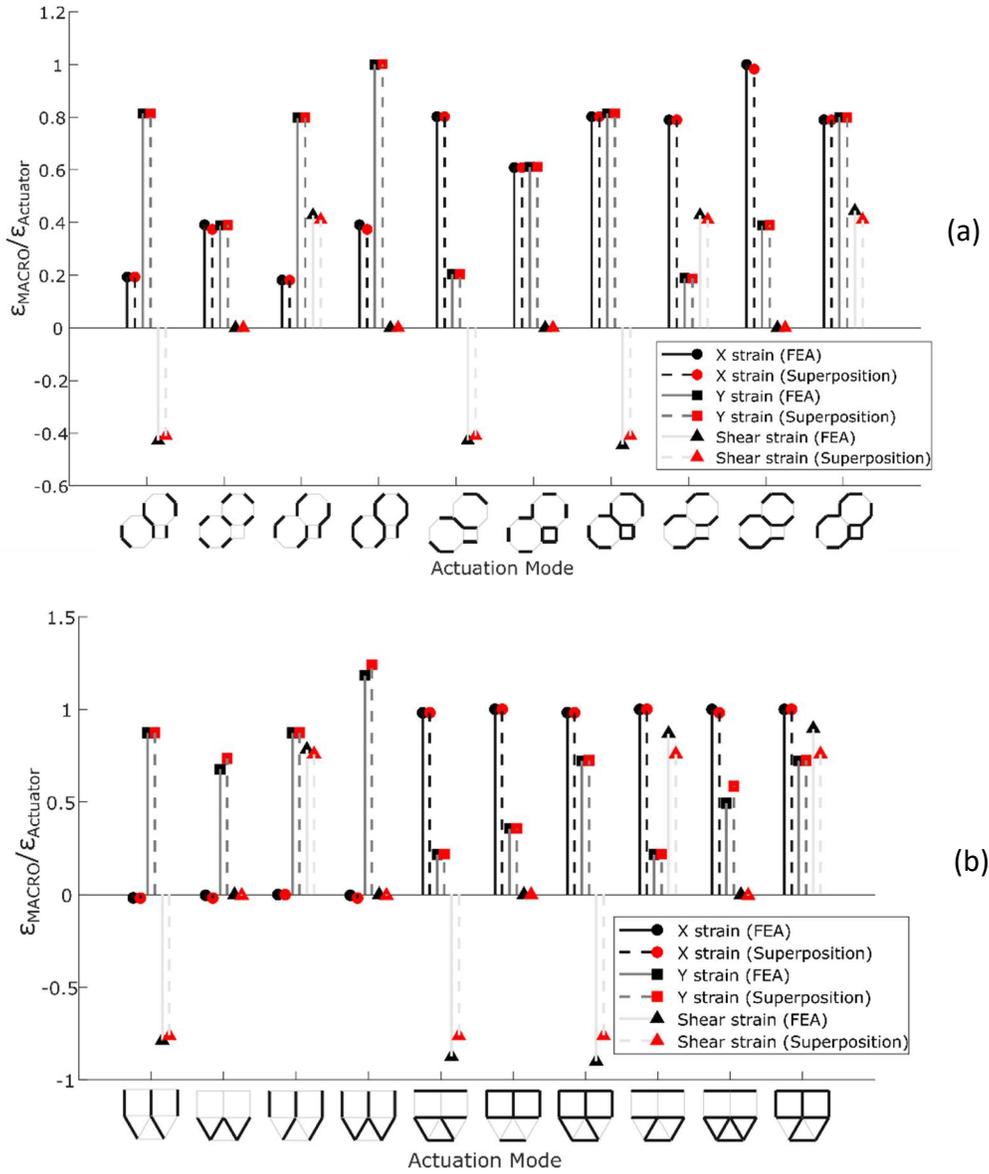

Figure 9. Comparison of strain ratios obtained from FEA with strain ratios obtained by adding the base deformations for (a) $T_3S_2$ and (b) $SO_2$ mesh.

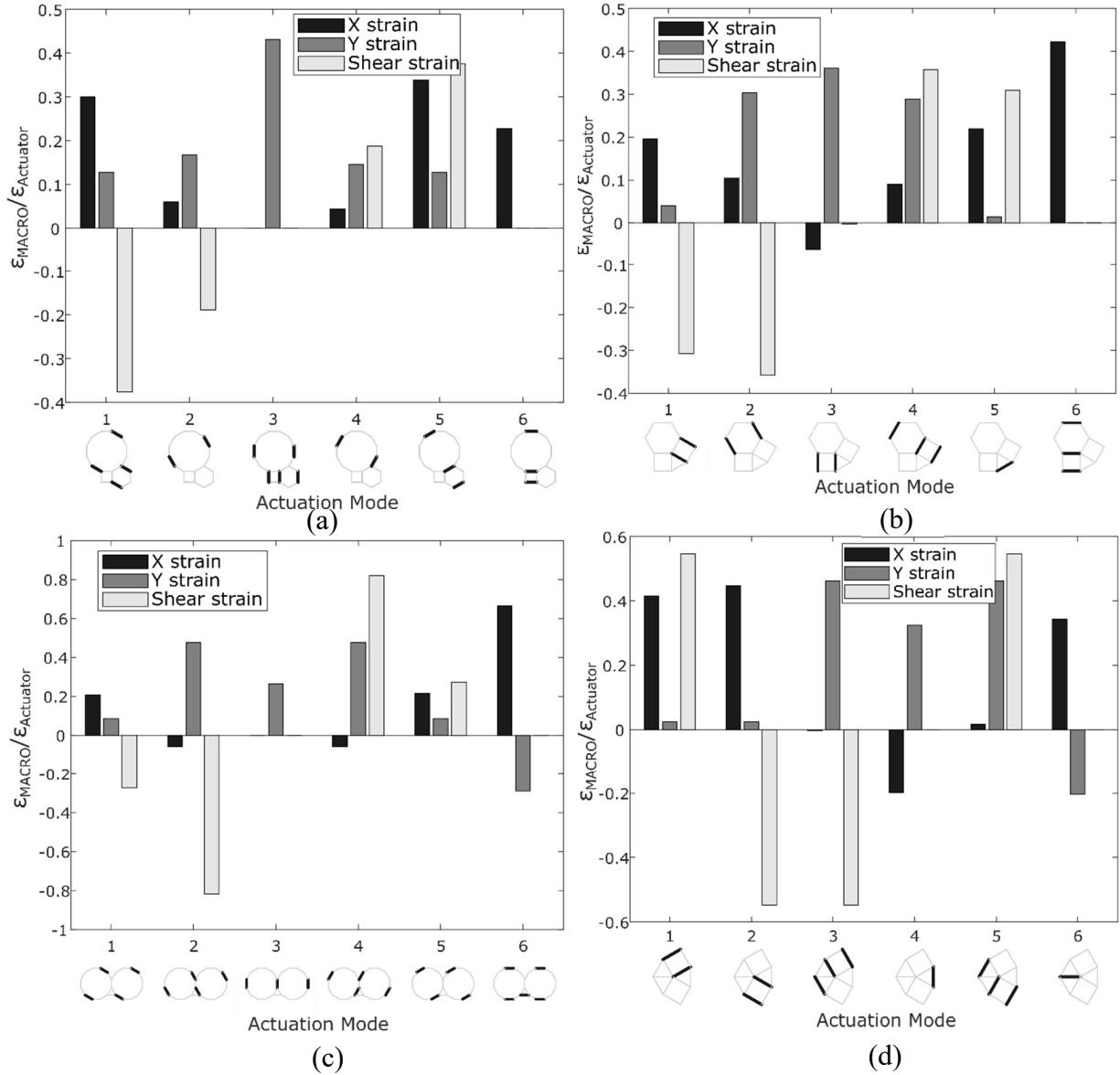

Figure 10. Spanning set of actuation modes of (a) SHD, (b) TSHS, (c) TD$_2$, and (d) T$_2$STS mesh topologies.

### 3.3 Actuation Effort

In this subsection, we look at the energy required to actuate different mesh topologies considered in this paper. Since we have simulated different MACRO meshes only under internal actuation of its edges, the strain energy of the mesh upon actuation would equal the energy required to actuate it. Therefore, we can compare the strain energy of different meshes as a measure of the ease of actuation [9]. Larger strain energy values indicate that a higher actuation effort is needed to actuate and vice versa. Strain energy for all the mesh topologies on meshes of identical size is shown in Figure 11. The numerical value here is the average value across all the actuation modes of a given mesh topology. For reference, we have plotted the strain energy corresponding to each actuation mode as scatter points on top of the bar plot.

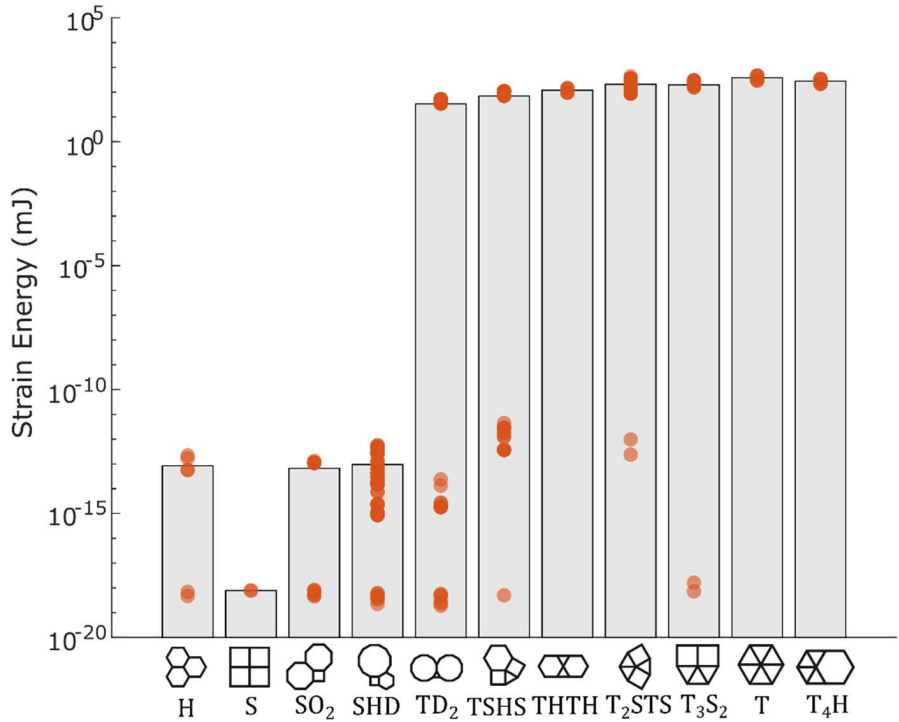

Figure 11. Comparison of the average of the strain energy of all actuation modes for each mesh topology. Strain energies corresponding to specific actuation modes for all mesh topologies are plotted as scatter points.

Figure 12 shows a heat map of the strain energies for the four different cases of varying relative actuator-node stiffness. The first column corresponds to the default case in which the actuators are stiffer and is the same data shown in Figure 11.

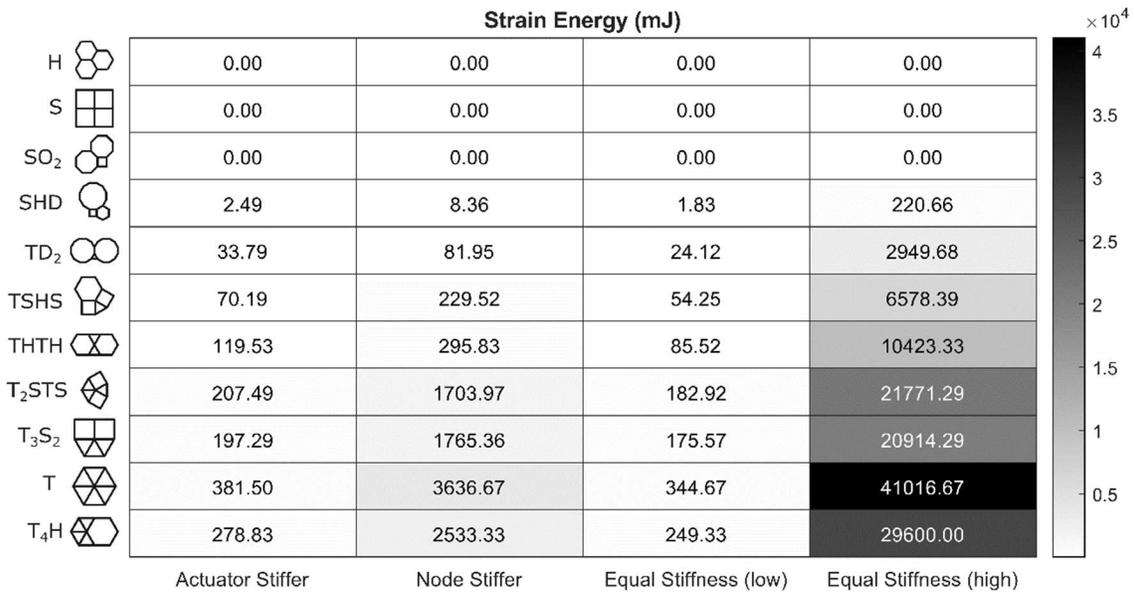

Figure 12. Heat map of strain energies for all mesh topologies (rows) and four different cases of relative actuator-node stiffnesses (column).

## 4. DISCUSSION

In this paper, we have characterized the strain behavior of the 11 possible regular planar meshes of MACROs. We simulated these MACRO meshes under different actuation modes that are based on the angular position of the edges in the mesh. For each mesh topology, out of the set of all such possible actuation modes, we identified a subset, $V$, that spans this whole set. We have shown these in Figure 6 through Figure 10 and these plots can be used to identify the actuation mode for a desired strain behavior, without running an FEA simulation.

Superposition of elements of $V$ can generate strains corresponding to all the possible actuation modes. It can also be shown that adding all the elements of $V$ would give the trivial case of all the edges being actuated with strain ratios being equal to one along both $X$ and $Y$ directions and zero shear strain. In other words, the overall mesh strain equals the individual actuator strain.

The different mesh topologies are also characterized for the actuation effort using strain energy as a metric. From Figure 11, we can see that the topologies that have polygons other than triangles in them (H, S, SHD, and $SO_2$) have negligible strain energy upon actuation compared to remaining meshes. This is due to the presence of polygons other than triangles in these meshes, which leads to bending-dominated behavior resulting in lower strain energies. Among the meshes with finite amount of strain energy, the order follows the nodal connectivity of the mesh i.e., the number of edges connected to a node. Hence, T mesh with nodal connectivity of six has the highest strain energy followed by in order, $T_4H$, $T_3S_2$, and $T_2STS$, which all have a nodal connectivity of five. This is followed by THTH and TSHS with nodal connectivity of four and finally $TD_2$ with three.

## 5. CONCLUSIONS AND FUTURE WORK

We modelled and characterized the strain and stiffness behavior of different networks of identical linear actuators and compliant nodes in this paper. Under the MACRO framework, we considered uniform tilings of plane as the design topologies. We demonstrated that an actuation scheme based on the angular orientation of edges of the mesh is independent of both the size of the mesh and the input strain in the actuated edges. Among the possible actuation modes, a small subset of base actuation modes is identified for each mesh topology. Further, we show that the remaining actuation modes can be obtained by superposition of the base actuation modes.

We also compared all the different mesh topologies based on the energy required for actuation. We show that the rank order based on this metric follows the nodal connectivity with higher nodal connectivity leading to higher energy required for actuation. The T mesh requires the highest energy among all mesh topologies. Next, we evaluated and compared the coefficients of the elastic stiffness tensor for all topologies. Only T mesh has all high stiffness coefficients whereas other meshes have high stiffness only for certain loading direction and not for all. Therefore, an ideal choice of mesh topology would require a consideration of the trade-off between desired stiffness and actuation effort.

In future work, we plan to use the stiffness and strain characteristics obtained in this two-part paper to predict the deformation of MACROs that use a combination of different mesh topologies. The strain characteristics of different mesh topologies corresponding to their range of actuation modes can be used in the future to design a MACRO to match a target deformation. Similarly, combining different mesh topologies and by tailoring the nodal stiffness by varying its cross-section area can be used to design a MACRO with desired stiffness properties. In other words, solving the inverse problem of obtaining mesh topology and actuator/nodal geometry for a target strain and stiffness behavior. Additionally, we plan to develop physical prototypes of MACROs using commercially available linear actuators. We plan to use the physical prototypes to validate the results presented in this paper. Further, we also plan to extend the FE modelling and mechanical characterization to spatial (3D) MACROs.

**FUNDING**

This work was supported by the US National Science Foundation under grant 1832795, "EFRI C3 SoRo: Muscle-like Cellular Architectures and Compliant, Distributed Sensing and Control for Soft Robots".

**Figure Captions List**

Figure 1. Deformation of Triangular MACRO mesh using FEA. (a) A 28x18 mesh with 60% edges randomly selected to be actuated (Thick edges correspond to the actuated edges (ON) while thin edges correspond to unactuated edges (OFF)) and (b) corresponding deformed shape obtained using FEA. (c) Same triangular mesh but with edges at 0° and 120° (with respect to horizontal measured counter-clockwise) being actuated and (d) corresponding deformed shape. (e) All edges actuated and (f) corresponding deformed shape.

Figure 2. Schematic of a MACRO mesh with boundary conditions used in the FEA model. The mesh is simply supported along the bottom edge to constrain the rigid body motions. To evaluate the strains, a minimal bounding parallelogram is used, shown by dotted lines.

Figure 3. Flow chart showing the step-by-step procedure for evaluating the strains corresponding to different choice of actuated edges of a given MACRO mesh.

Figure 4. Ratio of overall mesh strain to applied actuator strain for different percentages of edges actuated in the MACRO mesh measured along (a) horizontal direction (X axis) and (b) vertical direction (Y axis).

Figure 5. All possible actuation modes for (a) hexagon (H) and (b) $SO_2$ topologies based on the angular orientation-based actuation scheme. Thick edges correspond to the actuated edges (ON) and thin edges correspond to unactuated edges (OFF).

Figure 6. Ratio of MACRO mesh strain to applied actuator strain corresponding to each actuation mode for (a) Square (S), (b) Triangle (T), and (c) Hexagon (H) topologies. The X-annotations denote the respective actuation modes, where edges that are thicker correspond to actuators that are ON and the remaining edges (thin) correspond to actuators being OFF.

Figure 7. Comparison of strains obtained from FEA with the strains obtained by adding the base deformations for (a) Triangle and (b) Hexagon mesh.

Figure 8. Spanning set of actuation modes of (a) $T_3S_2$ and (b) $SO_2$ mesh topologies.

Figure 9. Comparison of strain ratios obtained from FEA with strain ratios obtained by adding the base deformations for (a) $T_3S_2$ and (b) $SO_2$ mesh.

Figure 10. Spanning set of actuation modes of (a) SHD, (b) TSHS, (c) $TD_2$, and (d) $T_2STS$ mesh topologies.

Figure 11. Comparison of the average of the strain energy of all actuation modes for each mesh topology. Strain energies corresponding to specific actuation modes for all mesh topologies are plotted as scatter points.

Figure 12. Heat map of strain energies for all mesh topologies (rows) and four different cases of relative actuator-node stiffnesses (column).

**Table Caption List**

Table 1. Number of distinct angular orientations for edges and corresponding actuation modes for all mesh topologies